\documentclass[runningheads]{llncs}
\usepackage[T1]{fontenc}
\usepackage{amsmath}
\usepackage{amssymb}
\usepackage{graphicx,verbatim}
\usepackage{booktabs} 
\usepackage{multirow} 
\usepackage{threeparttable}
\usepackage{algorithm}
\usepackage{algorithmic}
\usepackage{xcolor}
\usepackage{subcaption}

\usepackage{cases}
\usepackage{hyperref}
\usepackage{color}

\hypersetup{colorlinks=true,
            linkcolor=blue,
            anchorcolor=blue,
            citecolor=blue}

\newcommand{\INPUT}{\REQUIRE}

\begin{document}
\title{Zero-shot Low-Field MRI Enhancement via Diffusion-Based Adaptive Contrast Transport}
%

\author{
Muyu Liu\inst{1} \and 
Chenhe Du\inst{1} \and 
Xuanyu Tian\inst{1} \and 
Qing Wu\inst{1} \and 
Xiao Wang\inst{1} \and 
Haonan Zhang\inst{2} \and 
Hongjiang Wei\inst{2} \and 
Yuyao Zhang\inst{1}
}

\authorrunning{M. Liu et al.}

\institute{
School of Information Science and Technology, ShanghaiTech University, Shanghai, China \and
School of Biomedical Engineering, Shanghai Jiao Tong University, Shanghai, China
\\
\email{zhangyy8@shanghaitech.edu.cn} 
}
  
\maketitle              
%

\begin{abstract}
Low-field (LF) magnetic resonance imaging (MRI) democratizes access to diagnostic imaging but is fundamentally limited by low signal-to-noise ratio and significant tissue contrast distortion due to field-dependent relaxation dynamics. Reconstructing high-field (HF) quality images from LF data is a blind inverse problem, severely challenged by the scarcity of paired training data and the unknown, non-linear contrast transformation operator. Existing zero-shot methods, which assume simplified linear degradation, often fail to recover authentic tissue contrast. In this paper, we propose \textbf{DACT} (\textbf{D}iffusion-Based \textbf{A}daptive \textbf{C}ontrast \textbf{T}ransport), a novel zero-shot framework that restores HF-quality images without paired supervision. DACT synergizes a pre-trained HF diffusion prior to ensure anatomical fidelity with a physically-informed adaptive forward model. Specifically, we introduce a differentiable Sinkhorn optimal transport module that explicitly models and corrects the intensity distribution shift between LF and HF domains during the reverse diffusion process. This allows the framework to dynamically learn the intractable contrast mapping while preserving topological consistency. Extensive experiments on simulated and real clinical LF datasets demonstrate that DACT achieves state-of-the-art performance, yielding reconstructions with superior structural detail and correct tissue contrast.

\keywords{Low-Field MRI  \and  Diffusion Models \and Optimal Transport.}
\end{abstract}

\section{Introduction}
Magnetic resonance imaging (MRI) is a cornerstone of modern diagnostics, providing unparalleled soft-tissue contrast without ionizing radiation. 
However, its high cost and stringent infrastructural requirements limit its accessibility, particularly in resource-limited settings. 
The advent of low-field (LF) MRI systems (strengths $< 1\text{T}$) promises to democratize this technology, enabling affordable, point-of-care imaging at the patient's bedside~\cite{marques2019low,sheth2021assessment}. Despite this potential, the clinical utility of LF MRI is fundamentally hampered by two physical limitations: a low signal-to-noise ratio (SNR) and, more critically, significant alterations in tissue contrast~\cite{campbell2019opportunities}. The goal of computational enhancement is therefore not just to denoise the image, but to restore a high-field (HF) equivalent in both structural fidelity and diagnostic contrast.

The primary challenge lies in the complex physics of tissue relaxation. The longitudinal ($T_1$) and transverse ($T_2$) relaxation times of tissues are not physical constants; they are dependent on the main magnetic field strength ($B_0$)~\cite{bottomley1984review,rooney2007magnetic}. As the field strength decreases, these relaxation times shift, causing a non-linear compression and transformation of the image's intensity distribution. This results in muted distinctions between tissue types---such as gray and white matter---potentially compromising diagnostic accuracy. Consequently, enhancing LF MRI is a blind inverse problem where the forward degradation operator, specifically its contrast transformation component, is unknown and analytically intractable. This distinguishes the task from standard image restoration problems like denoising or super-resolution.

To address this, deep learning-based methods have been explored~\cite{iglesias2022quantitative,lin2019deep,tapp2024super}. Supervised approaches, which learn a direct mapping from LF to HF images, are severely hampered by the data scarcity dilemma; acquiring perfectly aligned, paired scans from the same subject on different scanners is logistically and ethically challenging ~\cite{man2023deep,zhao2024whole}. 
This has motivated a shift towards zero-shot methods that leverage powerful generative priors, such as Diffusion Models~\cite{chung2022diffusion,du2024dper,song2021solving}. However, the efficacy of these methods hinges on the accuracy of the forward model. In the context of LF MRI, existing frameworks fall into two pitfalls: they either erroneously model the degradation as purely spatial (ignoring contrast shifts entirely) ~\cite{lin2024zero}, or approximate the complex, non-linear relaxation dynamics via unconstrained operators~\cite{fei2023generative,gou2023test}. These physically insufficient assumptions fail to capture the monotonic yet non-linear nature of tissue signal changes, resulting in reconstructions that either retain the washed-out low-field contrast or suffer from intensity hallucinations.

To bridge this critical gap, we introduce \textbf{DACT} (\textbf{D}iffusion-Based \textbf{A}daptive \textbf{C}ontrast \textbf{T}ransport), a novel zero-shot framework designed to solve the blind inverse problem of LF MRI enhancement without paired supervision. DACT establishes a powerful synergy between a pre-trained diffusion model, which serves as a potent anatomical prior, and a novel, physically-informed adaptive forward model. Our core innovation is a differentiable contrast transport module that leverages Sinkhorn optimal transport~\cite{cuturi2013sinkhorn}. Guided by the physical assumption that the relative ordering of tissue intensities remains topologically consistent across field strengths~\cite{rooney2007magnetic}, this module dynamically estimates and corrects the non-linear intensity distribution shift during the reconstruction process. This allows DACT to learn the appropriate contrast mapping for each specific image instance, bypassing the need for an explicit, predefined forward model.

Our contributions are threefold:
(1) We propose DACT, the first zero-shot framework to address the unknown, non-linear contrast transformation inherent in LF MRI enhancement.
(2) We introduce a differentiable, physically-informed adaptive contrast transport module based on Sinkhorn optimal transport, which dynamically corrects for intensity distribution shifts within the reverse diffusion process.
(3) We demonstrate through extensive experiments on both simulated and real-world clinical datasets that DACT achieves state-of-the-art (SOTA) reconstruction quality.

\section{Methodology}



\begin{figure}[t]
\includegraphics[width=\textwidth]{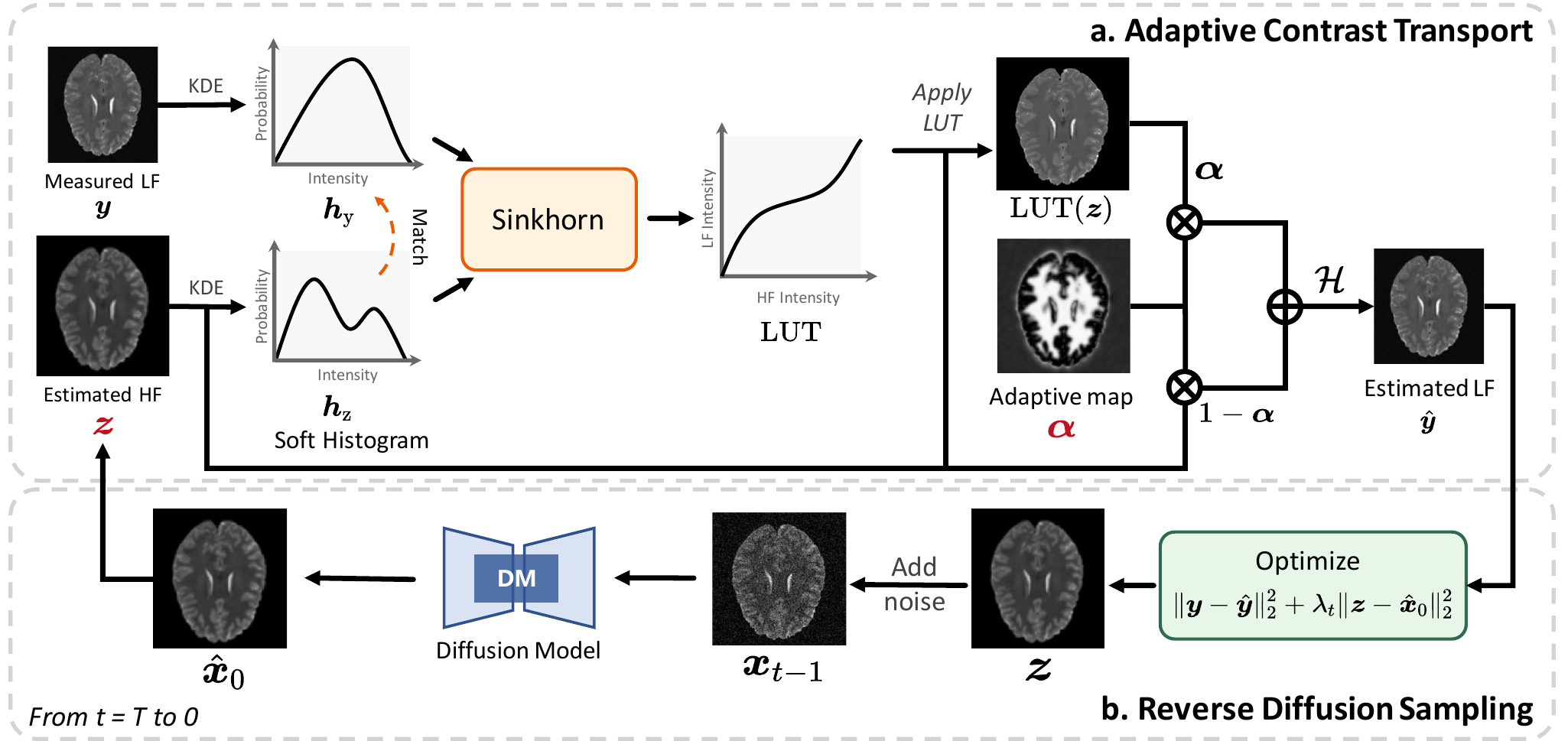}
\caption{\textbf{Overview of the DACT}.
(\textbf{a}) Soft histograms of input LF image $\boldsymbol{y}$ and HF estimate $\boldsymbol{z}$ are generated via kernel density estimation (KDE). The Sinkhorn algorithm computes an optimal transport plan for contrast alignment, while a learnable weight map $\boldsymbol{\alpha}$ adaptively balances contrast transformation and structural detail preservation. 
The known spatial degradation $\mathcal{H}$ is then applied to obtain the reconstructed LF estimate $\hat{\boldsymbol{y}}$.
(\textbf{b}) 
A high-field prior $\hat{\boldsymbol{x}}_0$ generated by a pretrained diffusion model is incorporated to jointly optimize the target HF image and the forward degradation process.
}
\label{figs:pipeline}
\end{figure}

We aim to reconstruct high-field (HF) equivalent images from low-field (LF) images in an unsupervised manner, and propose diffusion-based adaptive contrast transport (DACT). 
In DACT, we formulate the unknown HF-to-LF degradation as an \textit{adaptive contrast transport} and solve it via differentiable histogram mapping and the Sinkhorn algorithm~\cite{cuturi2013sinkhorn} (Sec.~\ref{sec:2.2}).  To regularize the ill-posed problem, a pre-trained diffusion model is leveraged to provide a generative HF prior~\cite{ho2020denoising,song2020score}, effectively constraining the solution space during optimization (Sec.~\ref{sec:2.3}). Fig~\ref{figs:pipeline} shows the pipeline of DACT.

\subsection{Problem Formulation}

The HF-to-LF degradation process is formulated as:
\begin{equation}
    \boldsymbol{y} = \mathcal{H}\boldsymbol{\Phi}(\boldsymbol{x}) + \boldsymbol{n},
    \label{eq:forward_model}
\end{equation} 
where $\boldsymbol{x}$ and $\boldsymbol{y}$ represent the HF and LF MRI, respectively. $\mathcal{H}$ denotes linear spatial downsampling and $\boldsymbol{\Phi}(\cdot)$ is the non-linear contrast transformation. 
In practice, the transformation $\boldsymbol{\Phi}$  is typically \textit{unknown and analytically intractable}, as tissue relaxation properties exhibit complex, non-linear variations across different magnetic field strengths~\cite{rooney2007magnetic}. Consequently, reconstructing an HF-equivalent image from an LF observation is a severely ill-posed problem, necessitating both accurate degradation modeling and the integration of effective data priors.  

\subsection{Degradation Modeling via Adaptive Contrast Transport}
\label{sec:2.2}
Given the complexity and intractability of the physical process $\boldsymbol{\Phi}$, we characterize the unknown degradation via \textbf{optimal transport (OT)}  between LF and HF intensity distributions.  
This formulation is advantageous for two reasons. First, it operates purely in the intensity space, decoupling contrast variations from spatial deformations.
Second, the integration of the \textbf{Sinkhorn algorithm}~\cite{cuturi2013sinkhorn} ensures a monotonic transport map—a crucial inductive bias that prevents intensity inversion and maintains the relative order of tissue intensities. 
Thus, estimating $\boldsymbol{\Phi}$ becomes an OT-based histogram alignment between the estimated HF and the LF image.
Fig.~\ref{figs:pipeline} (\textbf{a}) demonstrates the degradation modeling. 


\subsubsection{Monotonic Differentiable Histogram Matching.}
Standard histogram matching is inherently discrete and non-differentiable, posing a significant challenge for integration into gradient-based optimization frameworks. 
To overcome this, we employ the Sinkhorn algorithm to formulate a monotonic and differentiable histogram matching scheme.  
This approach ensures seamless compatibility with end-to-end optimization while simultaneously enforcing crucial physical constraints through its inherent monotonicity.

Specifically, we first employ kernel density estimation (KDE) to approximate the intensity distribution of both LF image and HF estimate to generate soft histograms $\boldsymbol{h}_{\text{LF}}$ and $\boldsymbol{h}_{\text{HF}}$. 
Then, we use the Sinkhorn algorithm to compute the optimal transport plan between these histograms, using a squared Euclidean cost to theoretically ensure monotonicity~\cite{brenier1991polar}. 
This plan is subsequently condensed into a smooth 1D look-up table (\textbf{LUT}) by calculating the conditional expectation of target intensities. 
Finally, the contrast transformation is realized by applying the LUT to $\boldsymbol{x}$ through differentiable linear interpolation.

\subsubsection{Adaptive Contrast Transport.}
While the derived LUT effectively captures the global non-linear contrast shift, strictly enforcing a 1D statistical match can be overly aggressive, potentially over-smoothing local high-frequency details. To mitigate this, we propose an adaptive operator $\hat{\boldsymbol{\Phi}}(\boldsymbol{x})$ as a learnable \textit{pixel-wise} convex combination of the LUT-transformed image and the original input:
\begin{equation}
    \hat{\boldsymbol{\Phi}}(\boldsymbol{x}; \boldsymbol{\alpha}) = \boldsymbol{\alpha} \odot \text{LUT}(\boldsymbol{x}) + (\mathbf{1}-\boldsymbol{\alpha}) \odot \boldsymbol{x},
\end{equation}
where $\odot$ denotes the element-wise product, and $\boldsymbol{\alpha} \in [0, 1]^{H \times W}$ is a spatially adaptive weight map. This tensor formulation allows the optimization to dynamically preserve HF structural details in textured regions ($\boldsymbol{\alpha} \to 0$) while enforcing strong contrast correction in homogeneous tissue areas ($\boldsymbol{\alpha} \to 1$).

\begin{algorithm}[t]
\small 
\caption{DACT: Diffusion-Based Adaptive Contrast Transport}
\label{alg:dact}
\begin{algorithmic}[1]
    \INPUT Pre-trained diffusion model $\mathcal{D}_{\boldsymbol{\theta}}$, measurement $\boldsymbol{y}$, spatial degradation $\mathcal{H}$, total timesteps $T$, inner iterations $J$, step sizes $\eta_{\boldsymbol{x}}, \eta_{\boldsymbol{\alpha}}$.
    \STATE \textbf{Initialize:} $\boldsymbol{x}_{T} \sim \mathcal{N}(\boldsymbol{0}, \boldsymbol{I})$.
    \STATE \textbf{Initialize:} $\boldsymbol{\alpha} \in \mathbb{R}^{H \times W}$ initialized to $\mathbf{0.5}$. \textcolor{gray}{// Initialize spatial weight map}
    
    \FOR{$t = T$ \textbf{to} $1$} 
         
        

        \STATE $\hat{\boldsymbol{x}}_{0} = \mathcal{D}_{\boldsymbol{\theta}}(\boldsymbol{x}_t; \sqrt{1/\rho_t})$ 
        \hfill \textcolor{gray}{\(\triangleright\) \textit{1. Diffusion Prior Injection}}
        
        \STATE $\boldsymbol{z} \leftarrow \hat{\boldsymbol{x}}_{0}$
        \hfill \textcolor{gray}{\(\triangleright\) \textit{2. Adaptive Data Fidelity Optimization}}
        \FOR {$j = 1$ \textbf{to} $J$}
            
            \STATE $\text{LUT} \leftarrow \text{Sinkhorn}(\boldsymbol{h}_{\boldsymbol{z}}, \boldsymbol{h}_{\boldsymbol{y}})$ 
            
            \STATE $\hat{\Phi}(\boldsymbol{z}; \boldsymbol{\alpha}) = \boldsymbol{\alpha} \odot \text{LUT}(\boldsymbol{z}) + (1-\boldsymbol{\alpha}) \odot \boldsymbol{z}$
            
            \STATE $\mathcal{L} = \|\boldsymbol{y} - \mathcal{H}\hat{\boldsymbol{\Phi}}(\boldsymbol{z}; \boldsymbol{\alpha})\|_2^2 + \lambda_t \|\boldsymbol{z} - \hat{\boldsymbol{x}}_{0}\|_2^2$

            \STATE $\boldsymbol{z} \leftarrow \boldsymbol{z} - \eta_{\boldsymbol{x}} \nabla_{\boldsymbol{z}} \mathcal{L}$ \quad $\boldsymbol{\alpha} \leftarrow \boldsymbol{\alpha} - \eta_{\boldsymbol{\alpha}} \nabla_{\boldsymbol{\alpha}} \mathcal{L}$ 
            \hfill \textcolor{gray}{// Joint Gradient Update}
            

        \ENDFOR
        
        

        \STATE  $\boldsymbol{x}_{t-1} = \text{DDIM}(\boldsymbol{z})$
            
    \ENDFOR
    \STATE \textbf{return} $\boldsymbol{x}_{0}$
\end{algorithmic}
\end{algorithm}

\subsection{Prior injection via Diffusion Model}
\label{sec:2.3}

Given that our proposed Sinkhorn-based adaptive contrast transport $\hat{\boldsymbol{\Phi}}(\boldsymbol{x})$ is fully differentiable, DACT can be seamlessly integrated into established diffusion-based inverse problem solvers. We adopt the widely used DiffPIR~\cite{zhu2023denoising} strategy, which solves the inverse problem via half-quadratic splitting (HQS)  based plug-and-play (PnP) framework~\cite{tian2025unsupervised,zhang2017learning}. 
Formally, the optimization objective can be formulated as:
\begin{equation}
    \hat{\boldsymbol{x}} = \underset{\boldsymbol{x}}{\arg\min} \|\boldsymbol{y} - \mathcal{H}\hat{\boldsymbol{\Phi}}(\boldsymbol{x})\|_2^2 + \lambda \mathcal{R}(\boldsymbol{x}),
\end{equation}
where $\mathcal{R}(\boldsymbol{x})$ represents the implicit prior provided by the pre-trained diffusion model. 
By introducing an auxiliary variable $\boldsymbol{z}$, the original problem is transformed to two subproblems:
\begin{subequations}
    \label{eq:m1.hqs}
    \begin{numcases}{}
        \boldsymbol{z}_k = \textrm{Denoiser}(\boldsymbol{x}_k; \sqrt{1/\rho_k}), \label{eq:m1.prior_sub}\\
        \boldsymbol{x}_k = \arg\min_{\boldsymbol{x}} \| \boldsymbol{y} - \mathcal{H}\hat{\boldsymbol{\Phi}}(\boldsymbol{x}; \boldsymbol{\alpha}) \|_2^2 + \rho_k \| \boldsymbol{x} - \boldsymbol{z}_{k-1} \|_2^2, \label{eq:m1.data_sub}
    \end{numcases}
\end{subequations}
where $\rho_k$ is a penalty parameter that controls the coupling strength between the data fidelity and the data prior. Through alternating optimization, these two subproblems synergistically resolve the ill-posed nature of LF MRI enhancement. Specifically, the data fidelity update (Eq.~\eqref{eq:m1.data_sub}) enforces consistency with the observed LF measurements, explicitly aligning both spatial topology and intensity distributions via our differentiable transport operator. Conversely, the prior update (Eq.~\eqref{eq:m1.prior_sub}) leverages the pre-trained diffusion model to seamlessly project the intermediate estimate back onto the natural HF MRI manifold. Ultimately, DACT can faithfully recover both authentic tissue contrast and high-frequency anatomical details from heavily degraded LF measurements. The overall optimization process is summarized in Algorithm~\ref{alg:dact}.

\section{Experiments \& Results}

\begin{table}[t!]
\centering

\caption{Quantitative comparisons of DACT with baselines on both the synthetic and real-world LF dataset. The best performances are highlighted in \textbf{bold}.}
\label{hcp}

\resizebox{\linewidth}{!}{
\begin{tabular}{ccccccccccc}
\toprule
\multirow{2.5}{*}{\textbf{Category}} & \multirow{2.5}{*}{\textbf{Method}} & \multicolumn{3}{c}{\textbf{Synthetic T1w}} & \multicolumn{3}{c}{\textbf{Synthetic T2w}} & \multicolumn{3}{c}{\textbf{Real T2w}}\\
\cmidrule(lr){3-5} \cmidrule(lr){6-8} \cmidrule(lr){9-11}
& & PSNR$\uparrow$ & SSIM$\uparrow$ & LPIPS$\downarrow$  & PSNR$\uparrow$ & SSIM$\uparrow$ & LPIPS$\downarrow$ & NIQE$\downarrow$ & BRISQUE$\downarrow$ & FID$\downarrow$ \\
\cmidrule{1-11}
\multirow{1}{*}{INR-based} 
& ULFINR  & 22.06 & 0.717 & 0.219 & 
            20.26 & 0.569 & 0.270 &
            6.614 & 29.53 & 169.65 \\
\cmidrule{1-11}
\multirow{1}{*}{Supervised} 
& PF-SR   & 21.92 & 0.705  & 0.213  &
            21.21  & 0.654  & 0.200  &
            5.452 & 24.18 &  118.37  \\
\cmidrule{1-11}
\multirow{4}{*}{\begin{tabular}[l]{@{}c@{}}Diffusion-\\based\end{tabular}}

& DiffDeuR    & 23.17 & 0.684 & 0.261 & 
            23.66 & 0.809 & 0.262  &
            5.887 & 20.67 & 130.12 \\

& GDP     & 23.42 & 0.679 & 0.217 & 
            23.92 & 0.766 & 0.228  &
            \textbf{4.821} & 24.45 & 115.96 \\

& TAO     & 23.61 & 0.689 & 0.216 & 
            24.18 & 0.782 & 0.209  &
            5.000 & 19.91 & 110.93 \\

& DACT (Ours)    & \textbf{27.47} & \textbf{0.832} & \textbf{0.140} & 
            \textbf{25.91} & \textbf{0.830} & \textbf{0.124}  &
            5.013 & \textbf{16.75} & \textbf{99.18} \\

\bottomrule
\end{tabular}
}
\end{table}
\begin{figure}[t!]
\centering
\includegraphics[width=\textwidth]{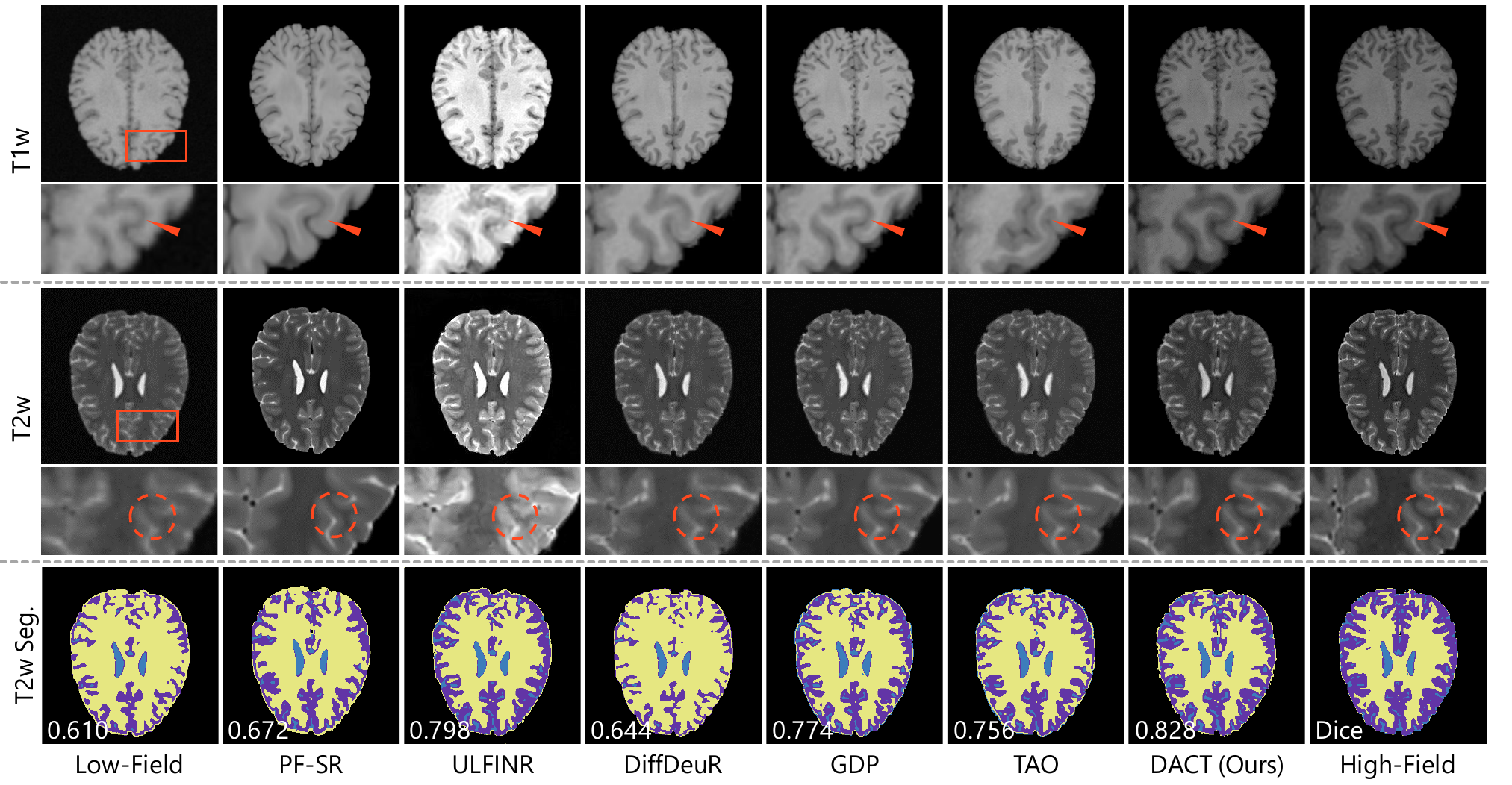}
\caption{Qualitative comparisons of DACT with baselines on two representative synthetic LF data. The bottom row shows the downstream segmentation masks for GM, WM, and CSF based on the T2w results, with the average Dice scores.} \label{figs:main-results-hcp}
\end{figure} 

\subsection{Experimental Setup}
\noindent \textbf{Datasets.}
(1) \underline{Synthetic Low-field data}: We utilize the HCP dataset~\cite{van2013wu} for simulation. Following the protocol in~\cite{lin2023low}, we approximate a 0.2T field strength contrast and apply 
$2\times$ in-plane downsampling to generate LF-HF pairs. The dataset is randomly partitioned into training, validation, and testing sets with a ratio of 70:15:15.
(2) \underline{Real World Low-field data}: The LF MRI was collected using a 0.2T MR scanner to obtain T2 weighted scans with a resolution of $1.5 \times 1.5~\text{mm}^2$ and a layer thickness of 9 mm. 

\noindent \textbf{Baselines.} We evaluate our method against six representative baselines across three categories: 
(1) \underline{Unsupervised INR-based}: \textbf{ULFINR}~\cite{islam2025ultra}
(2) \underline{Supervised}: \textbf{PF-SR}~\cite{man2023deep};
and (3) \underline{Diffusion-based}: \textbf{DiffDeuR}~\cite{lin2024zero}, \textbf{GDP}~\cite{fei2023generative}, and \textbf{TAO}~\cite{gou2023test}. Notably, PF-SR is trained using its original synthesis strategy to highlight the performance degradation caused by the contrast domain gap between synthetic data and real low-field acquisitions. All diffusion-based methods share the same pre-trained diffusion prior.

\noindent \textbf{Metrics.} Reconstruction fidelity on synthetic data is evaluated via peak signal-to-noise ratio (PSNR), structural similarity index (SSIM), and learned perceptual image patch similarity (LPIPS).
For real-world data, we adopt three no-reference quality metrics (NIQE~\cite{mittal2012making}, BRISQUE~\cite{mittal2012no}, FID~\cite{heusel2017gans}) due to the lack of ground truth.
To further assess the clinical utility of the enhanced images, we perform brain tissue segmentation and compute the Dice similarity coefficient for gray matter (GM), white matter (WM), and cerebrospinal fluid (CSF).

\noindent\textbf{Implementation Details.}
We trained an unconditional diffusion model using the high-field (3T) training partition of the HCP dataset. During inference, we employed DDIM~\cite{song2020denoising} sampling with 50 steps to accelerate the process. The internal test-time optimization was performed using the Adam optimizer with a learning rate of $1 \times 10^{-1}$ for a total of 25 iterations per slice.

\begin{figure}[t]
\centering
\includegraphics[width=\textwidth]{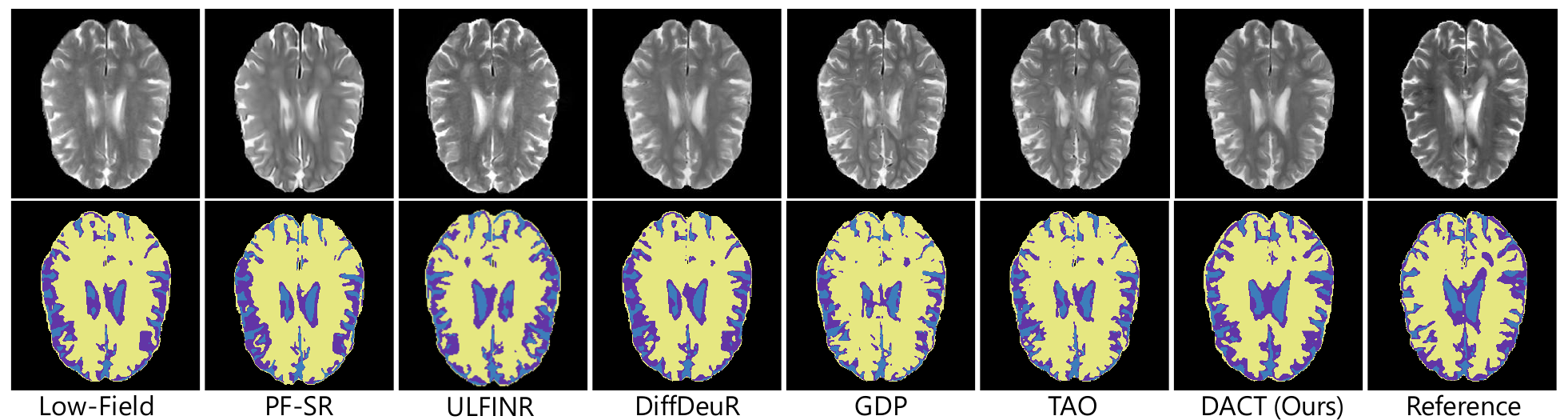}
\caption{Qualitative reconstruction (top) and tissue segmentation (bottom) on real LF data. DACT yields sharp cortical contrast and highly accurate segmentation.}
\label{figs:main-results-changhai}
\end{figure}

\subsection{Results}

\noindent\textbf{Results on Synthetic LF data.}
As shown in Table~\ref{hcp}, DACT consistently outperforms all competing methods across both T1w and T2w modalities. 
Visual comparisons in Fig.~\ref{figs:main-results-hcp} further validate our quantitative gains. Specifically, PF-SR suffers from a significant domain gap due to its data synthesis strategy, leading to over-blurred results. 
ULFINR fails to accurately reconstruct the characteristic high-field contrast. 
While diffusion-based methods (DiffDeuR, GDP, and TAO) attempt to restore details, they introduce noticeable anatomical deviations (highlighted by orange arrows and circles). 
In contrast, DACT effectively restores HF-equivalent contrast while preserving fine anatomical structures, demonstrating superior fidelity and high consistency with the ground truth.
Furthermore, the segmentation maps in Fig.~\ref{figs:main-results-hcp} further shows DACT’s enhanced clinical utility. 
Specifically, the GM segmentation derived from DACT
is markedly superior to baselines, and DACT achieves the optimal Dice score of 0.828.


\noindent\textbf{Results on Real-World LF data.}
Table~\ref{hcp} shows DACT achieves the best BRISQUE and FID scores. Qualitatively (Fig.~\ref{figs:main-results-changhai}), while baselines suffer from over-smoothing or artifacts, DACT robustly recovers sharp cortical folds and authentic contrast, enabling highly accurate downstream tissue segmentation.

\begin{figure}[t]
    \centering
    \begin{subfigure}[b]{0.45\textwidth}
        \centering
        \includegraphics[width=\textwidth]{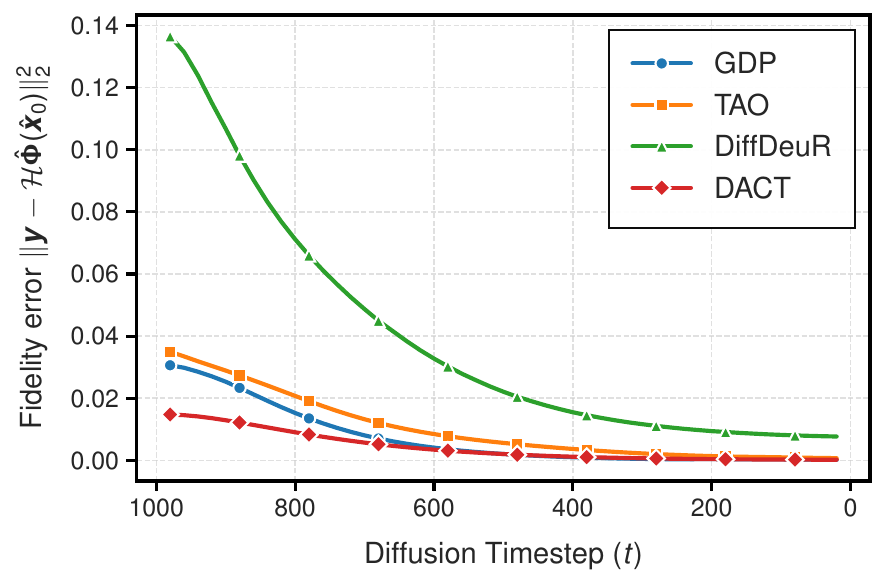}
    \end{subfigure}
    \begin{subfigure}[b]{0.45\textwidth}
        \centering
        \includegraphics[width=\textwidth]{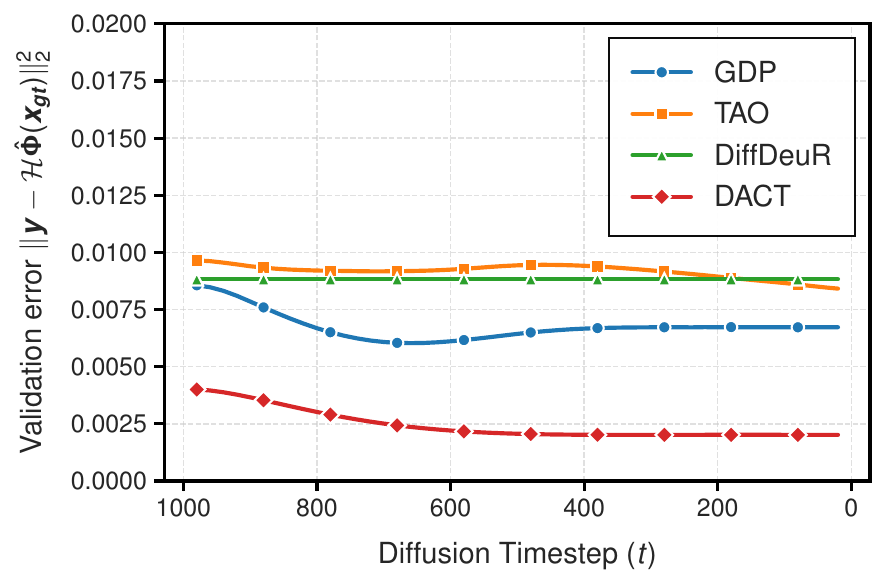}
    \end{subfigure}
    \caption{Convergence analysis. (Left) All methods successfully minimize fidelity error. (Right) Validation error against the ground truth reveals baseline overfitting, whereas DACT stably converges to the authentic contrast transformation.} 
    \label{figs:ablation}
\end{figure}

\subsection{Effectiveness of Adaptive Contrast Transport Modeling}
We analyze the convergence of various forward model formulations during the diffusion sampling process (Fig. \ref{figs:ablation}).
Specifically, DiffDeuR accounts only for spatial downsampling, GDP employs an affine contrast model ($f\boldsymbol{x}+m$), and TAO utilizes an unconstrained neural network.
Although the joint optimization loss decreases for all methods, evaluating their approximated forward models against the high-field ground truth ($\boldsymbol{x}_\text{gt}$) reveals that only DACT’s prediction error consistently decreases.
These results indicate that spatial-only modeling (DiffDeuR) is insufficient, while affine (GDP) or unconstrained (TAO) formulations fail to maintain physical consistency. 
In contrast, our Sinkhorn-based adaptive contrast transport effectively captures the unknown nonlinear degradation. Enforced monotonicity ensures a physically meaningful mapping, leading to stable convergence toward structurally and contrast-consistent reconstructions.

\section{Conclusion}

We present DACT, a physics-informed zero-shot framework for low-field MRI enhancement. To resolve unknown, non-linear contrast shifts, we introduce a Sinkhorn-based Adaptive Contrast Transport module. By leveraging differentiable optimal transport to align contrast distributions, DACT dynamically learns the non-linear intensity mapping while enforcing topological consistency and preventing hallucinated artifacts. Experiments on synthetic and real-world clinical datasets demonstrate state-of-the-art performance. Crucially, by faithfully recovering authentic tissue contrast and sharp anatomical details, DACT significantly benefits downstream tasks like tissue segmentation, offering a robust, data-efficient solution to bridge the low-field MRI domain gap.

\bibliographystyle{splncs04}
\bibliography{ref}

\end{document}